\documentclass[11pt]{article}

\usepackage[letterpaper, text={6.5in, 9.0in}]{geometry}
\usepackage{amsmath}
\usepackage{amssymb}
\usepackage{natbib}
\usepackage{graphicx}

\usepackage[mathlines]{lineno}

\def\bSigma{\mbox{\boldmath $\Sigma$}}

\def\bmu{\mbox{\boldmath $\mu$}}
\def\bxi{\mbox{\boldmath $\xi$}}

\def\bz{{\bf z}}

\def\bC{{\bf C}}
\def\bI{{\bf I}}

\def\bw{{\bf w}}

\def\bu{{\bf u}}
\def\bv{{\bf v}}
\def\bx{{\bf x}}

\def\bzero{{\bf 0}}

\def\mR{\mathbb{R}}
\def\bE{\mathbb{E}}

\begin{document}

\title{High-Dimensional Statistical Inference for Sparse Support Vector Machines}

\author{Peng Zeng\thanks{Corresponding author: Department of Mathematics and Statistics, Auburn University, AL 36849, United States. Email: zengpen@auburn.edu} \\
{\it Department of Mathematics and Statistics}\\
{\it Auburn University, Auburn, AL 36849}\\
zengpen@auburn.edu\\\\
Hanwen Huang \\
    {\it Department of Biostatistics, Data Science and Epidemiology}\\
    {\it Medical College of Georgia, Augusta University, Augusta, GA, 30912}\\
    hhuang1@augusta.edu
    }

\maketitle
\begin{abstract}
Using a replica-symmetric high-dimensional characterization, we develop an inferential framework for sparse support vector machines when the sample size and number of features grow proportionally. The main challenge is the nonsmooth hinge loss, which prevents direct application of debiasing arguments developed for smooth classification losses. We overcome this difficulty by representing the $L_1$-penalized support vector machine (SVM) as a linear program and identifying the hinge-loss subgradient through its dual variables. This yields a computationally accessible debiased estimator whose coordinates are asymptotically Gaussian under the proportional asymptotic regime. The resulting distributional characterization provides confidence intervals and hypothesis tests for individual features and enables false-discovery-rate-controlled variable selection. Extensive simulations examine calibration, power, and variable-selection performance under a range of covariance structures, including strongly correlated designs. An analysis of high-dimensional breast cancer gene-expression data illustrates how the proposed inference can distinguish statistically significant features from variables selected by the original sparse SVM.
\end{abstract}

\noindent\textbf{Keywords:} sparse support vector machine; high-dimensional inference; debiased estimator; nonsmooth loss; variable selection; proportional asymptotics

\section{Introduction}
The support vector machine (SVM) is a classical convex method for binary classification that remains useful in high-dimensional scientific applications in which prediction and feature interpretability are both important. In genomics and other molecular-data settings, for example, a classifier may be used not only to discriminate two biological states but also to identify a relatively small set of features for downstream investigation. This motivates sparse classification methods that combine predictive performance with interpretable variable selection. We do not view sparse SVMs as a universally dominant modern classifier; rather, they provide a particularly transparent convex benchmark in which sparsity, nonsmooth optimization, and high-dimensional inference interact in a mathematically tractable way.

A natural approach is to combine the SVM with an $L_1$ penalty. The resulting sparse SVM simultaneously performs classification and variable selection through the hinge-loss objective and the $L_1$ regularizer. This formulation has a long history \citep{bradley1998feature,zhu2004l1,zou2007improved}, and nonconvex penalties such as SCAD and MCP have also been considered to reduce regularization bias \citep{fan2001variable,WuYichao}. These developments have substantially advanced sparse classification for prediction and variable selection. However, rigorous statistical inference for sparse SVM coefficients remains much less developed in the proportional high-dimensional regime in which $n$ and $p$ grow at comparable rates.

The inferential problem is fundamentally different from the corresponding problem for smooth classification losses. Here, ``debiased'' refers to a bias-correction construction rather than unbiasedness in the classical finite-sample sense: the correction is designed to remove the first-order shrinkage induced by the $L_1$ penalty and to produce an estimator with an asymptotically Gaussian proxy. Thus, the relevant guarantee is asymptotic distributional equivalence, not the identity $\mathbb{E}(\bar{\bw})=\bw$ at finite sample size. In particular, the hinge loss
\[
V(u)=(1-u)_+,
\]
where $(z)_+:=\max(z,0)$ denotes the positive part of $z$, is nondifferentiable at the classification margin $u=1$. Consequently, score- and derivative-based debiasing arguments developed for smooth losses cannot be transferred directly to the SVM. The nonsmoothness must be handled both theoretically, in establishing the distributional behavior of the debiased estimator, and computationally, in evaluating the appropriate subgradient at the fitted estimator.

Classical asymptotic results for SVMs provide important foundations for inference. For example, \citet{koo2008bahadur} established a Bahadur representation for the linear SVM when the dimension is fixed. In the moderately high-dimensional regime, \citet{WuYichao} established an oracle property for nonconvex penalized SVMs and showed that a local linear approximation initialized at the $L_1$-penalized SVM can recover the oracle estimator under suitable conditions. Their results, however, do not provide coordinatewise distributional limits or formal confidence intervals and hypothesis tests in the proportional regime $p\asymp n$.

Recently, \citet{Huang_2025} developed a proportional-asymptotic inferential framework for $L_1$-penalized classification and demonstrated its effectiveness for logistic regression. The present work addresses the substantially different nonsmooth setting of the $L_1$-penalized SVM. Rather than treating the SVM as a routine specialization of the smooth-loss framework, we exploit its optimization structure: the $L_1$-penalized SVM admits an exact linear-program representation, and the LP dual variables identify a valid subgradient of the hinge loss. This correspondence provides the missing computational link between the nonsmooth optimization problem and the debiasing construction.

Our contributions are threefold. First, we establish a proportional-asymptotic characterization of the debiased $L_1$-penalized SVM under the Gaussian-mixture classification model, with its asymptotic behavior described through a finite-dimensional saddlepoint system. Second, we resolve the key nonsmoothness obstacle by showing that the hinge-loss subgradient can be recovered directly from the dual variables of the LP representation, leading to the computable correction
\[
\bar{\bw}
=
\hat{\bw}
+
\frac{1}{\zeta}\bSigma^{-1}\sum_{i=1}^n\nu_i y_i\bx_i.
\]
Here $\hat{\bw}$ is the fitted $L_1$-penalized SVM coefficient vector, $(y_i,\bx_i)_{i=1}^n$ is the observed sample of labels and feature vectors, $\bSigma$ is the feature covariance matrix, $\nu_i\ge0$ are the dual variables of the SVM's linear-program formulation identified in Section~\ref{section:implementation}, and $\zeta$ is a scalar calibration parameter obtained from the saddlepoint system introduced in Section~\ref{section:main-results}; formal definitions of all quantities appear in those two sections.
Third, the resulting Gaussian approximation yields coordinatewise confidence intervals and hypothesis tests and provides a basis for multiple-testing procedures and FDR-controlled variable selection. Numerical experiments examine calibration, power, and robustness under several correlation structures, and a breast cancer gene-expression analysis illustrates the distinction between variables selected by the sparse SVM and variables supported by formal statistical inference.

The present work is situated within a broader statistical-physics
literature on high-dimensional classification and generalized linear
models, of which \citet{Huang_2025} is the most closely related
instance. Other studies in this literature show that replica methods
can describe nonseparable perceptron regimes
\citep{Franz2019CriticalJammed} and class-imbalanced high-dimensional
linear classification \citep{Loffredo2024RestoringBalance}. These
studies help delimit the scope of our contribution: we develop
coefficient-level inference for the sparse hinge-loss SVM under the
balanced and linearly separable assumptions stated above.

The methodological novelty can therefore be summarized as
\[
\begin{gathered}
\text{proportional high-dimensional classification theory}\\
\;+\;\text{nonsmooth hinge-loss subgradient characterization}\\
\;+\;\text{LP-dual debiasing}\\
\;\longrightarrow\;\text{coefficient-level inference for sparse SVMs}.
\end{gathered}
\]
The proportional-asymptotic machinery in this decomposition is inherited from the existing high-dimensional classification framework of \citet{Huang_2025}; the new SVM-specific step is the treatment of the nonsmooth hinge loss through its LP-dual subgradient. This step is not a routine substitution of one loss function for another: the absence of a derivative at the margin requires a subgradient characterization that is compatible with both the asymptotic theory and the numerical optimization problem. The LP-dual representation supplies this characterization, so the debiasing correction is directly computable using quantities already returned by standard SVM optimization software.

The remainder of the paper is organized as follows. Section~\ref{section:main-results} introduces the high-dimensional SVM model, the debiased estimator, the Gaussian proxy, and the saddlepoint characterization, followed by the resulting inferential procedures. Section~\ref{section:implementation} develops the LP-dual implementation and establishes the connection between the dual variables and the hinge-loss subgradient. Section~\ref{section:numerical} evaluates the asymptotic approximation and inferential performance through simulations and a breast cancer gene-expression application. Section~\ref{section:conclusion} summarizes the contribution and discusses limitations and future directions.
\section{Asymptotic framework of high-dimensional classification}
\label{section:main-results}

In binary classification, we observe an independent sample $\{(y_i, \bx_i)\}_{i=1}^n$ from an unknown joint distribution $P(\bx, y)$, where $y_i \in \{-1, +1\}$ is the class label and $\bx_i \in \mathbb{R}^p$ is the covariate vector. The goal is to learn a linear decision rule $\operatorname{sign}(\bx^T\bw)$, parameterized by a weight vector $\bw \in \mathbb{R}^p$, that correctly predicts the label $y$ for any given $\bx$. The $L_1$-penalized SVM estimates $\bw$ by solving
\begin{align}\label{svm}
  \hat{\bw} = \operatorname*{arg\,min}_{\bw\in\mathbb{R}^p}
  \;\sum_{i=1}^n V\!\left(\frac{y_i\bx_i^T\bw}{\sqrt{p}}\right)
  + J_{\lambda}(\bw),
\end{align}
where $V(u) = (1-u)_+$ is the hinge loss and $J_{\lambda}(\bw)$ is the regularization term with tuning parameter $\lambda > 0$. The standard SVM uses $L_2$-regularization $J_{\lambda}(\bw) = \lambda\|\bw\|_2^2$; here we instead adopt $L_1$-regularization $J_{\lambda}(\bw) = \lambda\|\bw\|_1$, which promotes sparsity in $\hat\bw$ and enables variable selection.

We work within the proportional asymptotic framework of \citet{Huang_2025}, which takes the joint limit $n,p\to\infty$ with a fixed ratio $\alpha=n/p$. The data are assumed to arise from a symmetric two-component Gaussian mixture model:
\[
\bx\mid y=+1\sim N_p(\bmu,\bSigma),\qquad
\bx\mid y=-1\sim N_p(-\bmu,\bSigma),
\]
with balanced classes,
\[
\mathbb P(y=+1)=\mathbb P(y=-1)=\frac12.
\]
We assume balanced classes for analytical simplicity. High-dimensional
statistical-physics analyses of linear classifiers and SVMs under class
imbalance show that class proportions can affect classification
performance; see, for example, \citet{Loffredo2024RestoringBalance}.
Extending the present inferential framework to unequal class proportions
is an important direction for future work.
Here $\bmu\in\mathbb R^p$ denotes the mean-shift vector, $\bSigma$ is a positive definite covariance matrix, $\mu=\|\bmu\|$ is the signal strength, and $\hat{\bmu}=\bmu/\|\bmu\|$ is the unit signal direction. For any vector $\bv\in\mathbb R^p$, we define the $\bSigma$-weighted squared norm $\|\bv\|_{\bSigma}^2=\bv^\top\bSigma\bv$.

We further restrict attention to the \emph{linearly separable regime}, in which the fitted SVM admits a separating hyperplane for the training sample. This restriction is inherited from the asymptotic characterization used here and is important because the present derivation follows the replica-symmetric calculation for the separating solution. The restriction is not inherent to the replica method itself: related statistical-physics analyses have treated nonseparable (UNSAT or jammed) regimes of perceptron-type models; see, for example, \citet{Franz2019CriticalJammed}. Extending the present debiased-inference framework to the nonseparable SVM regime is beyond the scope of this work. We emphasize that the empirical optimization problem itself is convex: both the hinge loss and the $L_1$ penalty are convex, so there are no spurious local minima, although the objective need not be strictly convex and the optimizer need not be unique in every finite-sample realization.

The saddlepoint characterization below is obtained from the replica method applied to the high-dimensional convex optimization problem. We use the replica-symmetric (RS) ansatz. As in the underlying framework of \citet{Huang_2025}, we do not provide a rigorous proof of the RS ansatz or a separate Almeida--Thouless-type stability analysis. Accordingly, the resulting equations should be interpreted as a replica prediction, whose validity is assessed empirically in the numerical experiments. This distinction is important: the close agreement between simulations and the saddlepoint formulas provides strong evidence for the prediction, but it is not itself a proof of the replica calculation.

For completeness, we recall the notion of a subgradient. For a convex function $V$, a number $g$ is a subgradient of $V$ at $u$ if
\[
V(v)\ge V(u)+g(v-u)\qquad\text{for all }v.
\]
The set of all such $g$ is denoted $\partial V(u)$. For the hinge loss $V(u)=(1-u)_+$,
\[
\partial V(u)=
\begin{cases}
\{-1\}, & u<1,\\
[-1,0], & u=1,\\
\{0\}, & u>1.
\end{cases}
\]
Thus the subgradient is unique except exactly at the margin. The LP-dual construction in Section~\ref{section:implementation} selects a valid member of this set. Under the generic continuous Gaussian design considered here, the event that an observation lies exactly on the margin has probability zero; consequently the resulting subgradient, and hence the debiasing correction, is unique almost surely.

The inferential procedure is based on a debiased estimator $\bar{\bw}$ and a Gaussian proxy $\tilde{\bw}$. The key point is that, although the original SVM optimization problem is $p$-dimensional, its asymptotic behavior is characterized by a finite collection of scalar order parameters. This reduction makes the resulting inferential procedure computationally accessible in the proportional high-dimensional regime. The two quantities are defined as
\begin{align}
\bar{\bw} &= \hat{\bw} - \frac{1}{\sqrt{p}\,\zeta}\sum_{i=1}^n y_i\,\partial V\!\left(\frac{y_i\bx_i^T\hat{\bw}}{\sqrt{p}}\right)\bSigma^{-1}\bx_i, \label{df1}\\
\tilde{\bw} &= \frac{\sqrt{p}\,R_0}{\zeta}\,\bSigma^{-1}\hat{\bmu} + \tau\bSigma^{-1/2}\bz, \label{df2}
\end{align}
where $\partial V(\cdot)$ denotes the subgradient of the hinge loss, $\bz \sim N(\bzero, \bI_p)$, and $\tau = \sqrt{\zeta_0}/\zeta$. The scalar parameters $\zeta_0$, $\zeta$, $R_0$, $q_0$, $q$, and $R$ are the solution to the following saddlepoint system:
\begin{alignat}{3}
   \zeta_0 &= \frac{\alpha}{q^{2}}\,\bE\!\left(\hat{u}_\epsilon - R\mu - \sqrt{q_0}\,\epsilon\right)^{2}, &\qquad
   q_0 &= \frac{1}{p}\,\bE\|\hat{\bw}_{\bz}\|_{\bSigma}^2, & \label{nle}\\
   \zeta &= -\frac{\alpha}{q\sqrt{q_0}}\,\bE\!\left[(\hat{u}_\epsilon - R\mu - \sqrt{q_0}\,\epsilon)\,\epsilon\right], &\qquad
   q &= \frac{1}{p\sqrt{\zeta_0}}\,\bE\!\left(\hat{\bw}_{\bz}^T \bSigma^{1/2}\bz\right), \notag \\
   R_0 &= \frac{\alpha\mu}{q}\,\bE\!\left(\hat{u}_\epsilon - R\mu - \sqrt{q_0}\,\epsilon\right), &\qquad
   R &= \frac{1}{\sqrt{p}}\,\bE\!\left(\hat{\bw}_{\bz}^T\hat{\bmu}\right). \notag
\end{alignat}
The expectations in the left column are over $\epsilon \sim N(0,1)$; those in the right column are over $\bz\sim N(\bzero,\bI_p)$. The order parameters have direct geometric interpretations in the replica calculation. The quantity $q_0$ is the asymptotic $\bSigma$-weighted self-overlap (squared norm) of the effective estimator, while $R$ measures its overlap with the signal direction $\hat{\bmu}$. The auxiliary scalar $q$ controls the effective scalar optimization associated with the hinge loss, and $R_0$ is the corresponding signal-alignment coefficient in the Gaussian proxy. The quantities $\zeta_0$ and $\zeta$ describe the effective fluctuation and curvature/response scales of the high-dimensional optimization, with $\tau^2=\zeta_0/\zeta^2$ determining the coordinatewise Gaussian variance. In a replica interpretation, these quantities summarize self-overlaps, overlaps with the signal, and the effective response of the estimator to the random high-dimensional design, reducing the original $p$-dimensional problem to a finite-dimensional system. The auxiliary minimizers $\hat{u}_\epsilon$ and $\hat{\bw}_{\bz}$ appearing in (\ref{nle}) are defined by
\begin{align}
\hat{u}_\epsilon &= \operatorname*{arg\,min}_{u \in \mR}\left[V(u) + \frac{(u - R\mu - \sqrt{q_0}\,\epsilon)^2}{2q}\right], \label{hatu}\\
\hat{\bw}_{\bz} &= \operatorname*{arg\,min}_{\bw \in \mR^p}\left[\frac{\zeta}{2}\|\bw\|_{\bSigma}^2 - \left\langle\sqrt{\zeta_0}\,\bSigma^{1/2}\bz + \sqrt{p}\,R_0\hat{\bmu},\,\bw\right\rangle + \sum_{j=1}^p J_\lambda(w_j)\right]. \label{hatz}
\end{align}

The central asymptotic result of \citet{Huang_2025} establishes that $\bar{\bw}$ and $\tilde{\bw}$ have the same asymptotic distribution as $n,p\to\infty$ with $\alpha=n/p$ fixed. We specialize this result to the $L_1$-penalized SVM by taking $V$ to be the hinge loss and $J_\lambda$ to be the $L_1$ penalty. The additional contribution of the present work is to show how the nonsmooth loss can be handled exactly through the LP-dual representation developed in Section~\ref{section:implementation}.

\paragraph{Interpretation of the Gaussian proxy.}
The asymptotic statement should be read as follows: the raw SVM estimator $\hat{\bw}$ is the solution of the original penalized empirical optimization problem and is therefore the object used for prediction. The corrected estimator $\bar{\bw}$ is constructed from $\hat{\bw}$ by removing the leading $L_1$-induced shrinkage. The vector $\tilde{\bw}$ is not another estimator computed from the data; it is a theoretical Gaussian representation whose distribution approximates that of $\bar{\bw}$. Consequently, $\hat{\bw}$, $\bar{\bw}$, and $\tilde{\bw}$ have distinct roles:
\[
\boxed{\begin{gathered}
\hat{\bw}:\ \text{prediction and sparse optimization},\\[2pt]
\bar{\bw}:\ \text{inference after bias correction},\\[2pt]
\tilde{\bw}:\ \text{asymptotic Gaussian proxy}.
\end{gathered}}
\]

The resulting Gaussian proxy provides the principal inferential consequence of the theory. Since $\tilde{\bw}$ in (\ref{df2}) is Gaussian with componentwise variance $\tau^2=\zeta_0/\zeta^2$, each component of the debiased estimator is asymptotically normal with the corresponding variance. Thus, for each coordinate $j$, a Wald-type confidence interval and a $p$-value for $H_0:w_j=0$ can be constructed from $\bar w_j$ and $\tau^2$. These $p$-values can subsequently be adjusted using a false discovery rate procedure \citep{benjamini1995controlling}, providing a principled route from sparse classification to formal variable selection with error-rate control. In addition, the asymptotic classification accuracy is
\[
\Phi\!\left(R\|\bmu\|/\sqrt{q_0}\right),
\]
where $R$ and $q_0$ are obtained from the saddlepoint system (\ref{nle}). Thus the same finite-dimensional characterization simultaneously describes predictive performance and coordinatewise inferential uncertainty.

\section{SVM Implementation}
\label{section:implementation}

The empirical optimization problem is convex, although generally not strictly convex: the hinge loss and the $L_1$ penalty are both convex. Hence the optimization does not suffer from spurious local minima. The role of the replica calculation is therefore not to identify a local optimum, but to characterize the high-dimensional statistical behavior of the global convex optimum.

The main computational difficulty is to evaluate the hinge-loss subgradient at the fitted estimator. Because the hinge loss is nondifferentiable at the classification margin, numerical differentiation is neither natural nor necessary. The $L_1$-penalized SVM has an exact linear-program representation whose dual variables provide the required subgradient directly. Thus the theoretical debiasing correction can be computed from quantities already returned by a standard LP solver.

The $L_1$-penalized SVM can be written as
\begin{align*}
  V(u) = (1 - u)_+, \qquad
  J_{\lambda}(w) = \lambda |w|.
\end{align*}
The framework requires evaluating the subgradient of the loss function in place of its derivative, since the hinge loss is not differentiable at $u = 1$. In the following, we show that this subgradient can be identified with the dual variables of an associated linear programming problem.

The resulting optimization problem is the following linear program:
\begin{align*}
   \min_{\bxi, \bw, \bu}
   \sum_{i=1}^{n} \xi_{i} + \lambda\sum_{j=1}^{p} u_{j}
\end{align*}
subject to
\begin{align*}
   y_{i}\bx_{i}^{T}\bw \geq \sqrt{p}(1 - \xi_{i}), \quad
   -u_{j}\leq w_{j} \leq u_{j}, \quad
   \xi_{i} \geq 0, \quad  i = 1, \ldots, n, \\
   u_{j} \geq 0, \quad j = 1, \ldots, p.
\end{align*}
Introducing Lagrange multipliers $\nu_i$, $s_j$, $t_j$, $\psi_i$, $\rho_j$ for each constraint in turn, the Lagrangian is
\begin{align*}
  L &= \sum_{i=1}^{n} \xi_{i} + \lambda\sum_{j=1}^{p} u_{j}
   + \sum_{i=1}^{n} \nu_{i}\bigl(\sqrt{p}(1 - \xi_{i}) - y_{i}\bx_{i}^{T}\bw\bigr) \\
   &\quad {}
   + \sum_{j=1}^{p} s_{j}(-u_{j} - w_{j})
   + \sum_{j=1}^{p} t_{j}(w_{j} - u_{j})
   - \sum_{i=1}^{n} \psi_{i}\xi_{i}
   - \sum_{j=1}^{p} \rho_{j}u_{j},
\end{align*}
where $\nu_i \geq 0$ and $\psi_i \geq 0$ for $i = 1,\ldots,n$,
and $s_j \geq 0$, $t_j \geq 0$, and $\rho_j \geq 0$ for $j = 1, \ldots, p$.
The KKT stationarity conditions are
\begin{align*}
   \frac{\partial L}{\partial \xi_i}
   = 1 - \sqrt{p}\nu_i - \psi_i &= 0,
   \quad i = 1, \ldots, n, \\
   \frac{\partial L}{\partial w_{j}}
   = - \sum_{i=1}^{n} \nu_{i} y_{i}x_{ij}
   - s_{j} + t_{j} &= 0,
   \quad j = 1, \ldots, p,  \\
 \nu_{i}\bigl(\sqrt{p}(1 - \xi_{i}) - y_{i}\bx_{i}^{T}\bw\bigr) &= 0,
 \quad i = 1, \ldots, n,\\
 \psi_i\xi_i &= 0,
 \quad i = 1, \ldots, n.
\end{align*}
The first stationarity condition implies $0 \leq \nu_i \leq 1/\sqrt{p}$.
The three possible cases for $\xi_i$ are:
\begin{itemize}
\item
When $\xi_{i} > 0$, we have $\psi_{i} = 0$,
which further implies $\nu_{i} = 1/\sqrt{p}$ and
$\xi_{i} = 1 - y_{i}\bx_{i}^{T}\bw/\sqrt{p}$.
\item
When $\xi_{i} = 0$ and $y_{i}\bx_{i}^{T}\bw/\sqrt{p} > 1$,
we have $\nu_{i} = 0$ and $\psi_{i} = 1$.
\item
When $\xi_{i} = 0$ and $y_{i}\bx_{i}^{T}\bw/\sqrt{p} = 1$,
we have $\nu_{i} \in [0, 1/\sqrt{p}]$ and $\psi_{i} \in [0, 1]$.
\end{itemize}
These cases can be summarized as
\begin{align*}
  \nu_i = \begin{cases}
      1/\sqrt{p}, & 1 - y_i\bx_i^T\bw/\sqrt{p} > 0, \\
      [0, 1/\sqrt{p}], & 1 - y_i\bx_i^T\bw/\sqrt{p} = 0, \\
      0, & 1 - y_i\bx_i^T\bw/\sqrt{p} < 0.
  \end{cases}
\end{align*}
The second stationarity condition gives
\begin{align*}
  \sum_{i=1}^n \nu_iy_ix_{ij} + (s_j - t_j) = 0,
  \quad j = 1, \ldots, p.
\end{align*}
At the same time, the stationarity condition of the original SVM problem (\ref{svm}) requires that
\begin{align*}
  0\in \sum_{i=1}^n \partial V_i \cdot  \frac{y_i\bx_i}{\sqrt{p}} + \lambda \partial \|\bw\|_1,
\end{align*}
where $\partial V_i = \partial V(y_i\bx_i^T\bw/\sqrt{p})$.
Comparing the two conditions, we identify the subgradient as $\partial V_i = -\nu_{i}\sqrt{p}$, or more precisely,
\begin{align*}
 \partial V_i = -\nu_i\sqrt{p}
 = \begin{cases}
      -1, & 1 - y_i\bx_i^T\bw/\sqrt{p} > 0, \\
      [-1, 0], & 1 - y_i\bx_i^T\bw/\sqrt{p} = 0, \\
      0, & 1 - y_i\bx_i^T\bw/\sqrt{p} < 0,
  \end{cases}
\end{align*}
which is consistent with the definition of the subgradient of the hinge loss.
Hence, the debiased estimator (\ref{df1}) can be expressed in closed form as
\begin{align*}
    \bar \bw
    &= \hat \bw - \frac{1}{\sqrt{p}\zeta} \sum_{i=1}^{n}
    y_{i} \partial V\!\left(\frac{y_{i}\bx_{i}^{T}\hat \bw}{\sqrt{p}}\right) \bSigma^{-1}\bx_{i}
     = \hat \bw + \frac{1}{\zeta} \bSigma^{-1} \sum_{i=1}^{n} \nu_iy_{i}\bx_{i}.
\end{align*}
Most standard linear programming solvers return the dual variables $\nu_i$ automatically.
We solve the linear program using Gurobi (\verb"https://www.gurobi.com/"), where the dual variables are returned as \verb"pi" in the solver output.

\paragraph{Geometric interpretation.}
The sparse SVM chooses a separating direction by trading off margin violations against the total absolute size of the coefficient vector. The $L_1$ penalty therefore favors separating hyperplanes whose normal vectors concentrate their mass on a relatively small number of coordinate directions. When $\bSigma=\bI_p$, the signal direction $\hat{\bmu}$ provides the natural reference direction, while the $L_1$ penalty can rotate the fitted normal toward coordinate-aligned sparse directions. For nonisotropic $\bSigma$, the geometry is governed jointly by the signal direction and the covariance metric: directions with smaller variance can provide larger standardized separation, while the $L_1$ penalty favors sparse coordinate representations. The order parameter $R$ quantifies alignment with the signal direction, whereas $q_0$ quantifies the overall $\bSigma$-weighted size of the effective classifier. The debiasing step does not change the fitted separating rule used for prediction; rather, it corrects the coefficient-level shrinkage so that uncertainty about individual coordinates can be quantified.

\section{Numerical studies}
\label{section:numerical}

This section evaluates the theoretical and practical performance of the proposed inferential procedure. We focus on four questions. First, does the proportional-asymptotic characterization accurately predict classification performance? Second, are the resulting confidence intervals calibrated across different covariance structures and sparsity levels? Third, do the resulting tests provide useful power and variable-selection performance? Fourth, does the method remain effective under strong predictor correlation? We also illustrate the procedure using a high-dimensional breast cancer gene-expression dataset.

The SVM models are formulated and solved as linear programs using the R package \verb"gurobi", which provides an interface to the Gurobi optimizer (\verb"https://www.gurobi.com/"). The LP dual variables required for the debiasing correction are returned by the same optimization step; no additional optimization problem is required. The code used for the numerical studies is available on \verb"https://github.com/pzengauburn/svm-lasso/".


\subsection{Classification accuracy}

This subsection compares the empirical classification accuracies computed by Monte Carlo with those computed from the asymptotic result in Section~\ref{section:main-results}.
We set $p=1000$ and $n=p\alpha$, where $\alpha=0.5$. The class labels $y\in\{-1,+1\}$ are sampled independently with $\mathbb P(y=+1)=\mathbb P(y=-1)=1/2$, matching the balanced-mixture assumption used in the asymptotic characterization.
The predictor $\bx\mid y$ is simulated from $N_p(y\bmu, \bSigma)$ for various configurations of $\bmu$ and $\bSigma$. We choose $\bmu = a\bSigma \bw_0$, where $\bw_0$ is a sparse vector with components equal to 1 or 0 and sparsity level $s = 0.01, 0.05, 0.1$, and $a$ is selected to ensure that the length of $\bmu$ is 2. Set $\bSigma = \sigma^2\bC$, where $\sigma^2 = 2.0$. We consider four different correlation structures for $\bC$: (1) IID, $\bC = \bI_p$, the identity matrix;
(2) block diagonal, $\bC$ is a block diagonal matrix with blocks $\begin{pmatrix} 1 & 0.8 \\ 0.8 & 1 \end{pmatrix}$;
(3) AR(1), $\bC = (0.8^{|i-j|})$ representing an autoregressive model of order 1;
(4) banded, $\bC$ is a banded correlation matrix where the diagonal elements are $C_{ii} = 1$ and the off-diagonal elements are $C_{ij} = 0.4$ if $i \neq j$ and $|i - j| \leq 2$ and $C_{ij} = 0$ otherwise.

The classification accuracy is the probability of correctly classifying a new observation, defined as $P(\tilde y\tilde\bx^T\hat\bw > 0)$, where $\hat\bw$ is the coefficient vector estimated from the training data and $(\tilde y, \tilde\bx)$ denotes a new observation.
In practice, the classification accuracy is approximated by the proportion of correctly classified observations in a test dataset of size $n$, simulated using the same mechanism as the training data.
As $n, p \to \infty$, the asymptotic classification accuracy is given by $\Phi(R\|\bmu\|/\sqrt{q_0})$, where $\Phi(\cdot)$ denotes the CDF of the standard normal distribution, and $R$ and $q_0$ are saddlepoint parameters.

\begin{figure}[htb]
\begin{center}
\includegraphics[width = 0.45\textwidth]{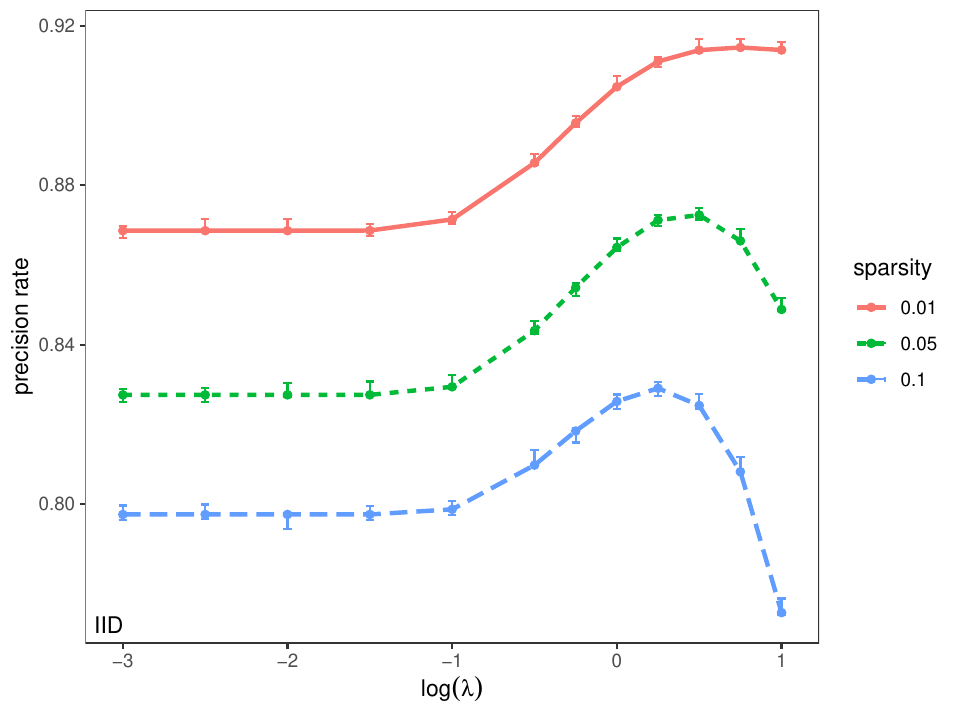}
\includegraphics[width = 0.45\textwidth]{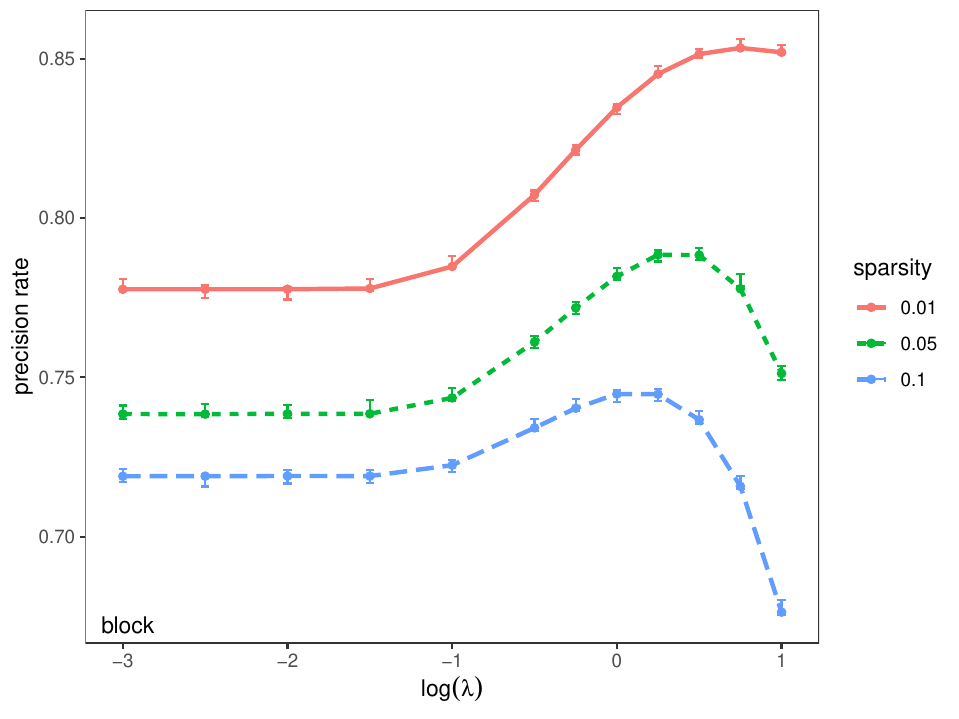}
\includegraphics[width = 0.45\textwidth]{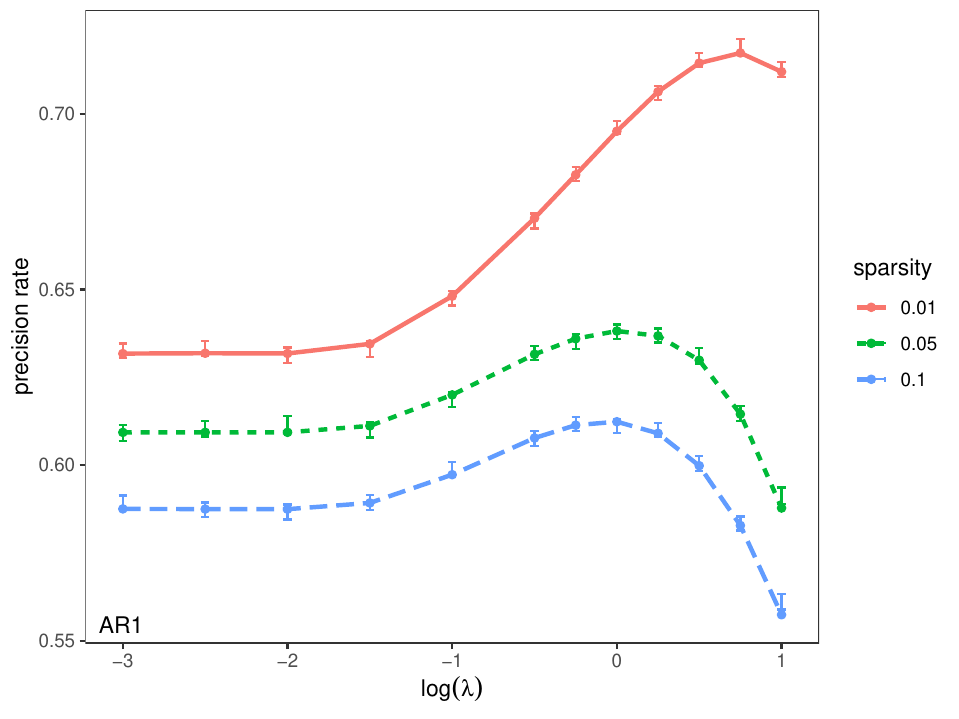}
\includegraphics[width = 0.45\textwidth]{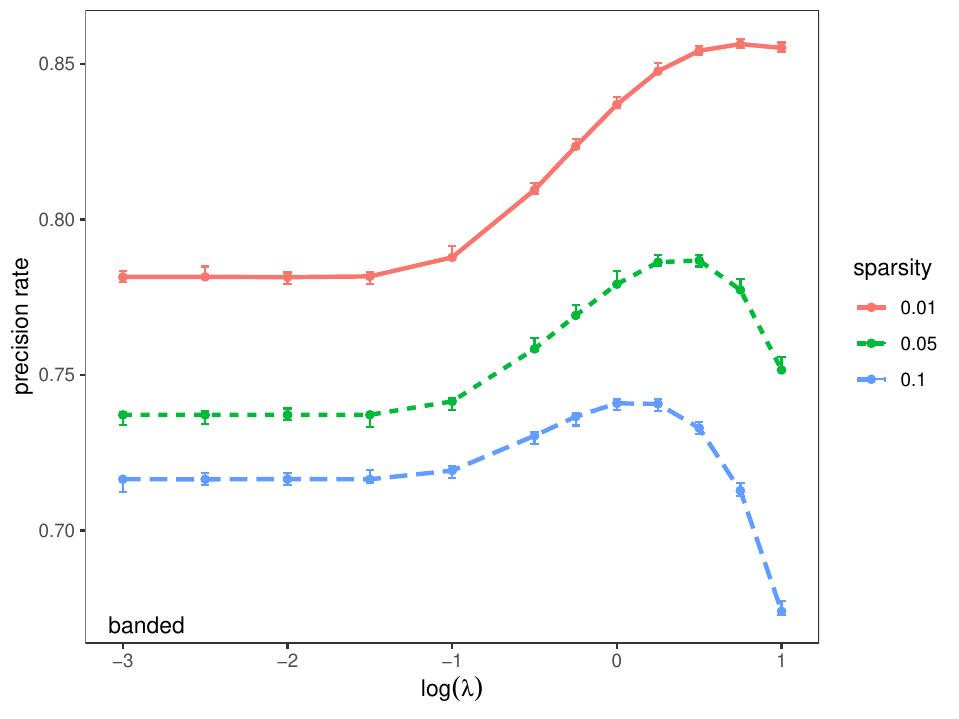}
\end{center}
\caption{Empirical and asymptotic classification accuracies as a function of $\log\lambda$ for four different correlation structures: IID (top-left), block (top-right), AR(1) (bottom-left), and banded (bottom-right). The lines are the asymptotic classification accuracies at different sparsity levels $s = 0.01, 0.05, 0.1$. The error bars are the 95\% confidence intervals for the mean classification accuracies based on 500 replicates.}
\label{fig:1-precision}
\end{figure}

Figure~\ref{fig:1-precision} compares the empirical and asymptotic classification accuracies under different settings. In each setting, we simulate 500 training datasets with a specified correlation structure and sparsity level. Each training dataset is fitted using a sequence of tuning parameters, $\log\lambda = -3, -2.5, \ldots, 1$.
The empirical classification accuracy is evaluated using separately simulated test data.
The error bars denote the 95\% confidence intervals for the mean classification accuracies, while the lines represent the asymptotic classification accuracies. The figure shows close agreement between the empirical and asymptotic rates.
The classification accuracy increases as $\bw_0$ becomes sparser for each correlation structure.
At each sparsity level, the classification accuracy decreases as the correlation structure becomes more complex. Specifically, the IID case yields the highest accuracy, whereas the AR(1) case results in the lowest accuracy.

\begin{figure}[htb]
\begin{center}
\includegraphics[width = 0.45\textwidth]{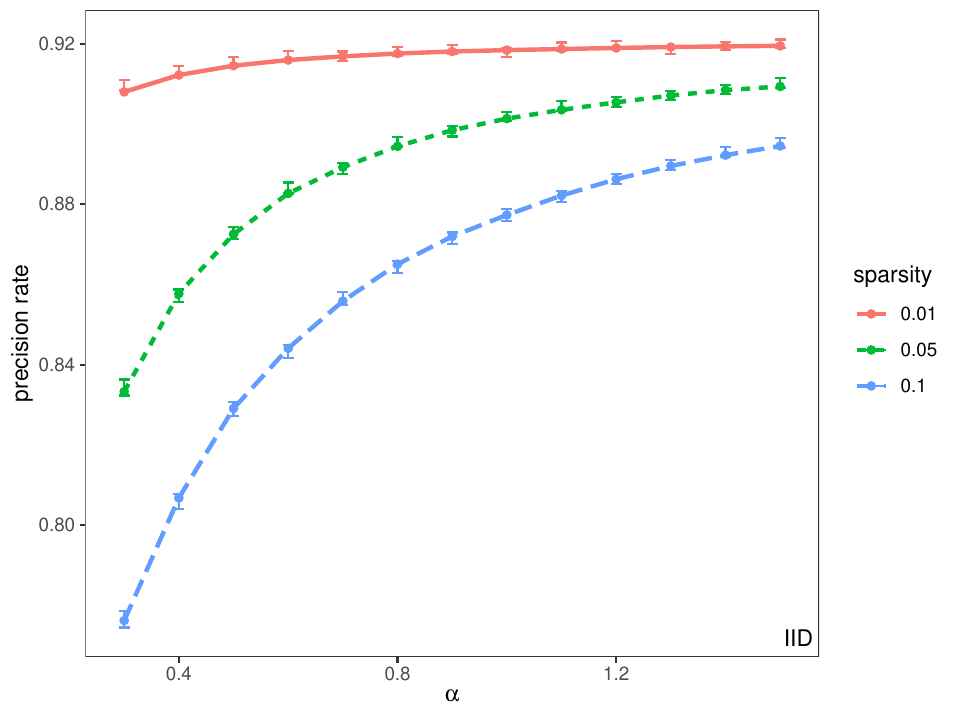}
\includegraphics[width = 0.45\textwidth]{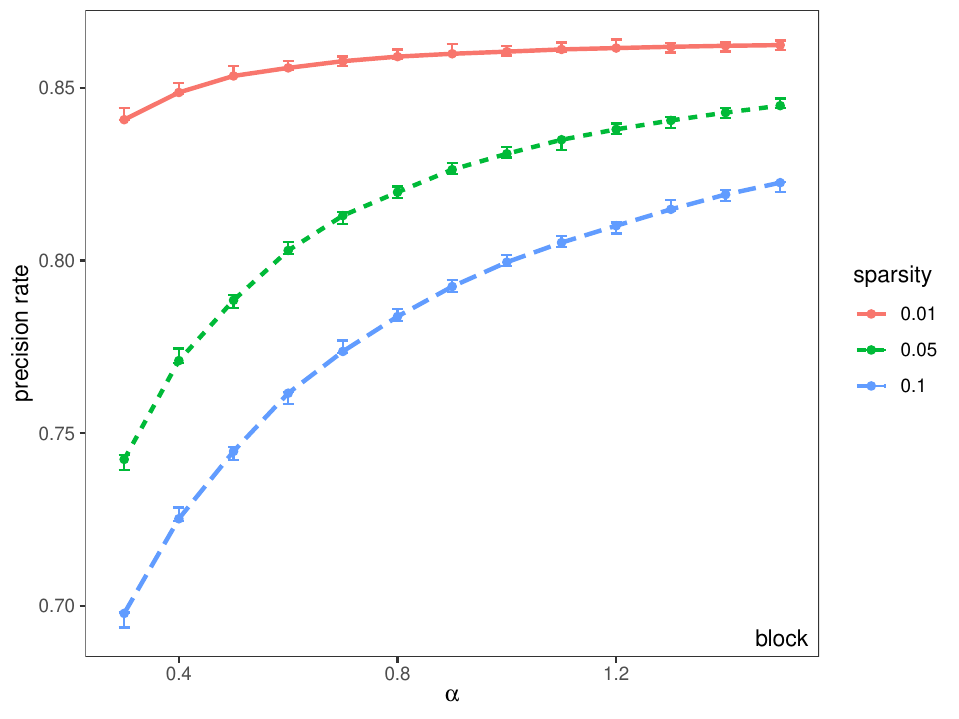}
\includegraphics[width = 0.45\textwidth]{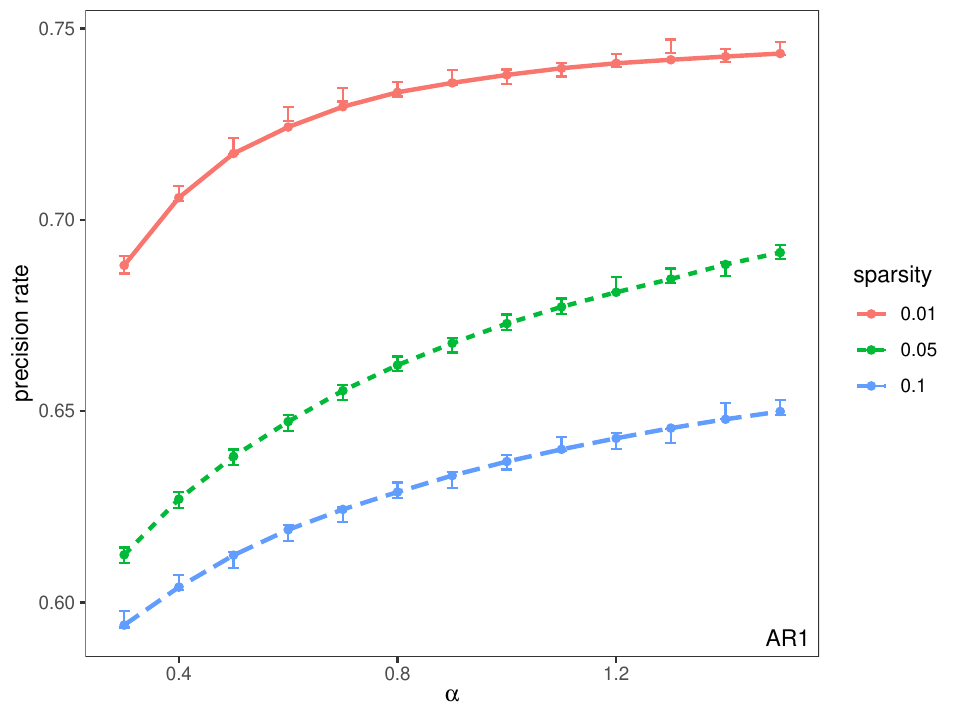}
\includegraphics[width = 0.45\textwidth]{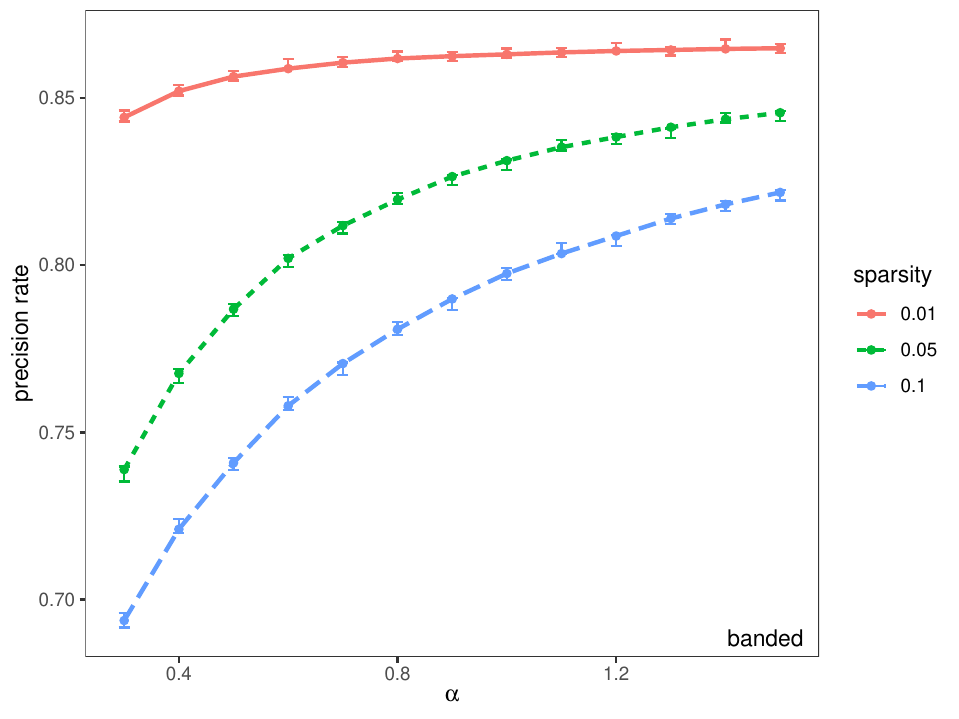}
\end{center}
\caption{Empirical and asymptotic optimal classification accuracies as a function of $\alpha$ for four different correlation structures: IID (top-left), block (top-right), AR(1) (bottom-left), and banded (bottom-right). The lines are the asymptotic optimal classification accuracies at different sparsity levels $s = 0.01, 0.05, 0.1$. The error bars are the 95\% confidence intervals for the mean optimal classification accuracies based on 500 replicates.}
\label{fig:2-optimal-precision}
\end{figure}

Figure~\ref{fig:1-precision} suggests that the classification accuracy depends on the choice of $\lambda$, where the optimal value of $\log\lambda$ lies approximately between $0$ and ${0.5}$, depending on the correlation structure and sparsity level.
To further investigate performance in terms of the optimal classification accuracy,
we conduct additional simulations for $\alpha = 0.3, \ldots, 1.5$, following a similar procedure, to examine how the optimal classification accuracy depends on $\alpha$.
The optimal classification accuracy is the best classification accuracy among the scenarios when $\log\lambda = -3, -2.5, \ldots, 1$.
In Figure~\ref{fig:2-optimal-precision}, the lines denote the asymptotic optimal classification accuracy at different sparsity levels, whereas the error bars represent the 95\% confidence intervals for the mean optimal classification accuracies.
The optimal classification accuracy increases as the true underlying model becomes sparser or as $\alpha$ increases. Both scenarios correspond to settings in which $n$ becomes much larger relative to the effective dimension, that is, the number of relevant variables.

\subsection{Variable selection}

This subsection evaluates the inferential calibration and variable-selection performance of the proposed procedure. We consider three aspects: asymptotic normality, confidence-interval coverage, and power for detecting nonzero coefficients. Because the proposed framework also produces coordinatewise $p$-values, we additionally describe how those $p$-values can be used for FDR-controlled selection.


Unlike the preceding subsection, which evaluates prediction using $\hat{\bw}$, the following experiments use $\bar{\bw}$ and its Gaussian proxy for inference. The purpose is to determine whether the correction removes the regularization-induced distortion sufficiently well to support calibrated uncertainty quantification.

\begin{figure}[htb]
\begin{center}
\includegraphics[width = 0.45\textwidth]{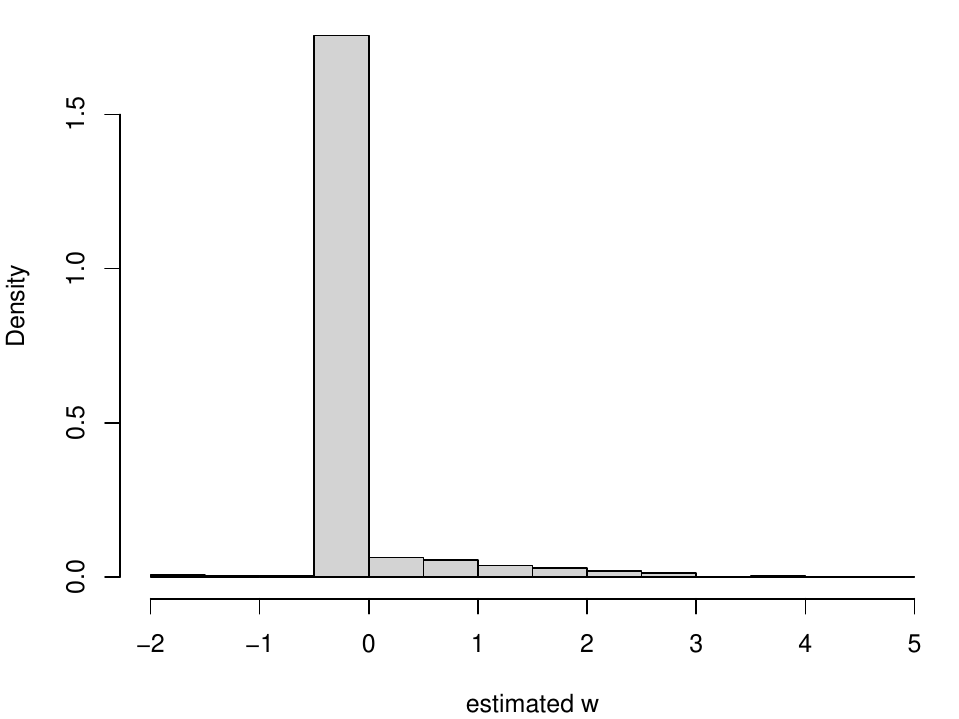}
\includegraphics[width = 0.45\textwidth]{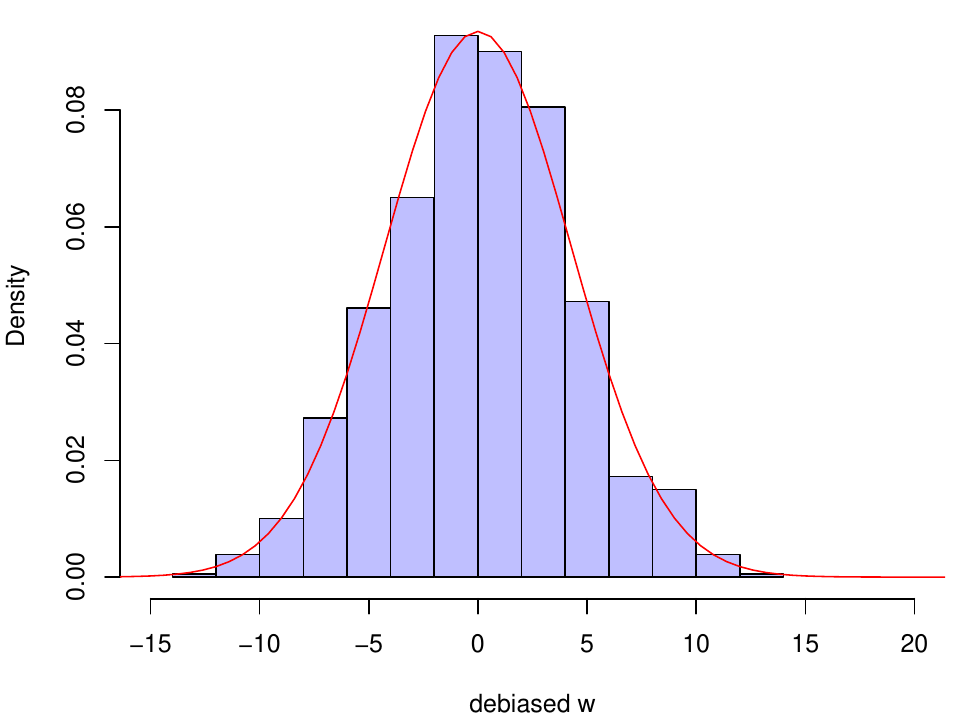}
\end{center}
\caption{The left plot shows the histogram of the estimated coefficients $\hat\bw$, whereas the right plot displays the histogram of the debiased estimates $\bar\bw$ for the zero components along with the asymptotic normal density curve.}
\label{fig:3-hist}
\end{figure}

First, we verify the asymptotic normality of the debiased estimator discussed in Section~\ref{section:main-results}.
We simulate a dataset using the AR(1) correlation structure with $\alpha = 0.5$ and $s = 0.1$, and fit the model using $\lambda = 1$. The left plot in Figure~\ref{fig:3-hist} shows the histogram of the estimated coefficients $\hat\bw$, which exhibits a large spike at zero, indicating that many estimated coefficients are exactly zero.
The right plot displays the histogram of the debiased estimates $\bar\bw$ for the zero components, along with the asymptotic normal density curve.
Note that the mean of $\bar\bw$ is $\sqrt{p}R_0\bSigma^{-1}\hat\bmu/\zeta$, corresponding to the first term in expression (\ref{df2}).
The asymptotic normal density curve closely matches the histogram.
Similar results are observed for other choices of correlation structure, $\alpha$, or $s$.

\begin{figure}[htb]
\begin{center}
\includegraphics[width = 0.45\textwidth]{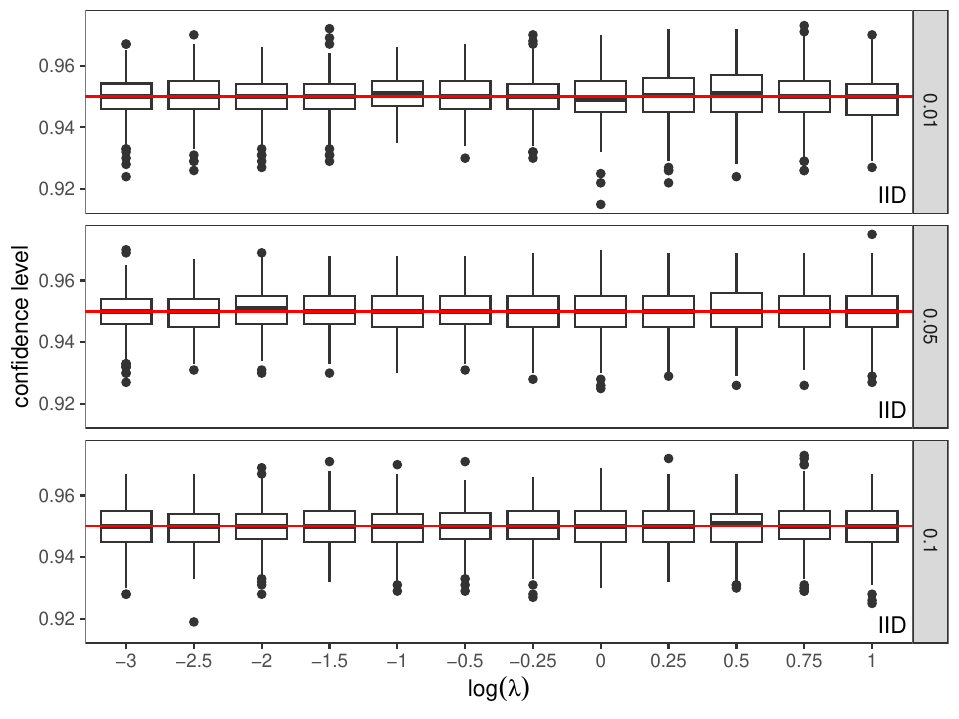}
\includegraphics[width = 0.45\textwidth]{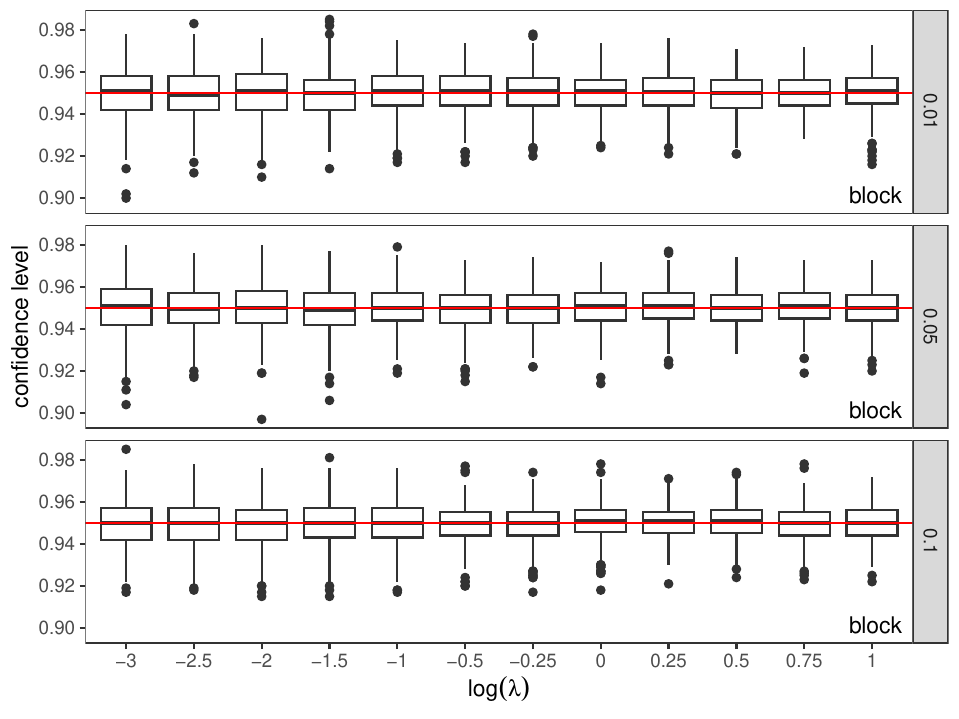}
\includegraphics[width = 0.45\textwidth]{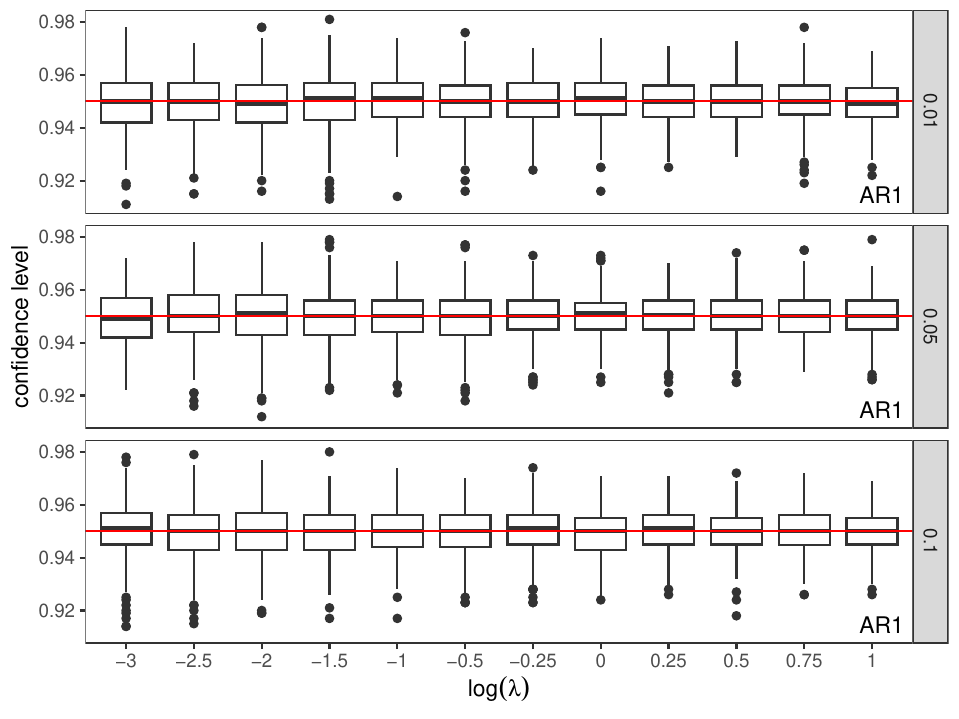}
\includegraphics[width = 0.45\textwidth]{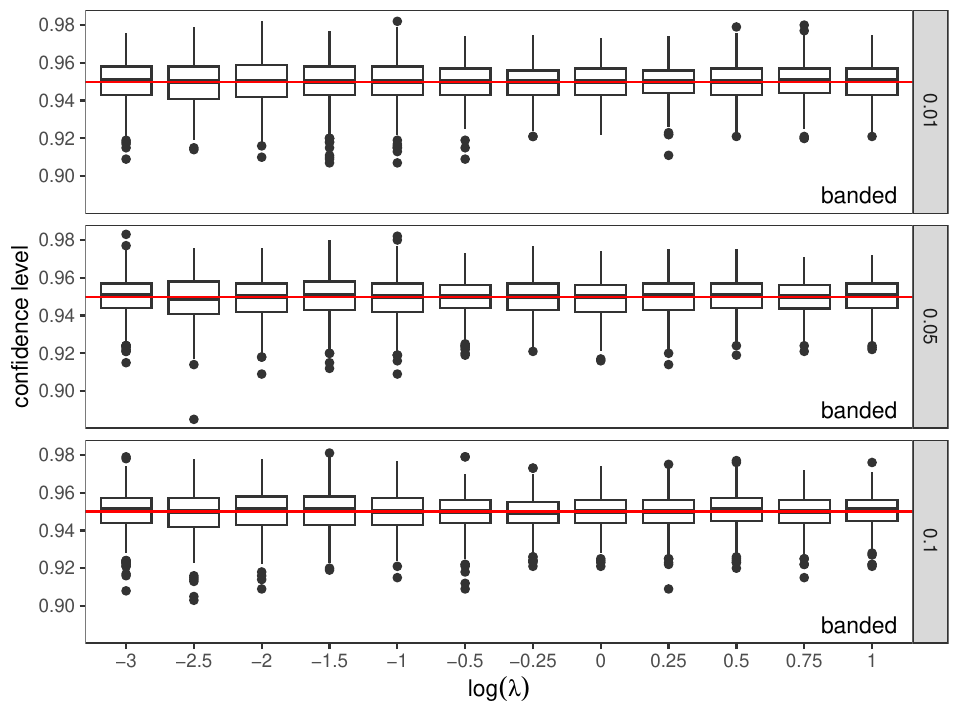}
\end{center}
\caption{Boxplots of empirical confidence levels based on 500 datasets for four correlation structures: IID (top-left), block (top-right), AR(1) (bottom-left), and banded (bottom-right). The horizontal line at 0.95 indicates the nominal confidence level. Different rows of plots represent different sparsity levels, $s = 0.01, 0.05, 0.1$.}
\label{fig:4-conf}
\end{figure}

Second, we validate the formula for constructing confidence intervals.
Recall that a 95\% confidence interval is expected to contain the true mean 95\% of the time when such intervals are constructed for datasets drawn from the same population.
We set $\alpha = 0.5$ and generate 500 datasets with a specified correlation structure and sparsity level. For each dataset, 95\% confidence intervals are constructed for every component.
The empirical confidence level is defined as the proportion of times these confidence intervals contain the corresponding true mean.
Figure~\ref{fig:4-conf} presents boxplots of the empirical confidence levels for each combination of correlation structure, sparsity level, and tuning parameter. The horizontal line at 0.95 represents the nominal confidence level. The close agreement across all settings confirms the validity of the confidence interval formula.

\begin{figure}[htb]
\begin{center}
\includegraphics[width = 0.45\textwidth]{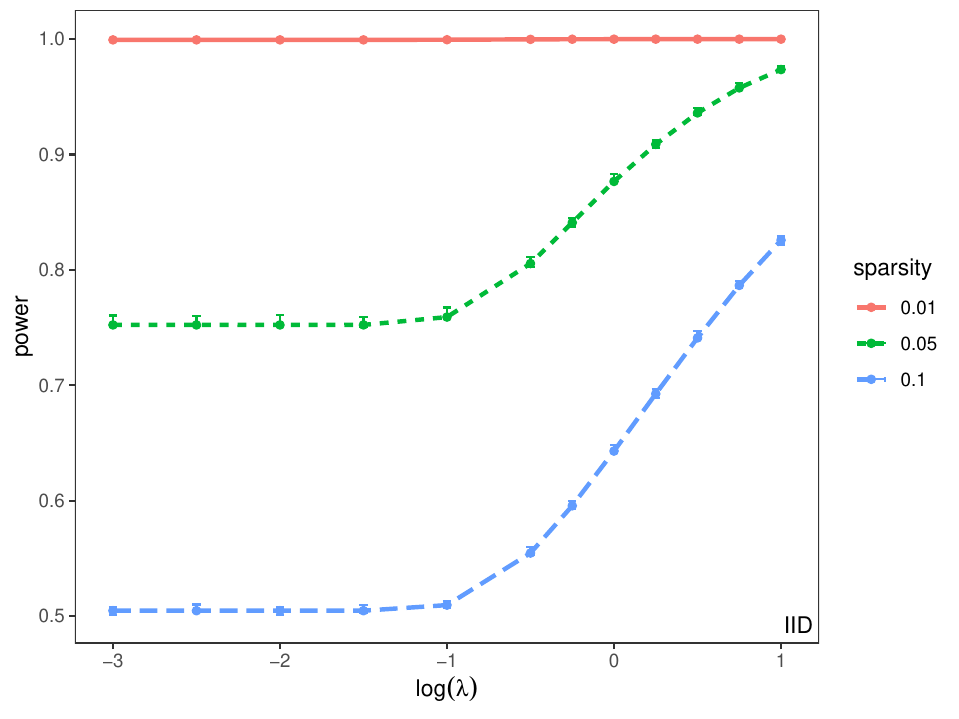}
\includegraphics[width = 0.45\textwidth]{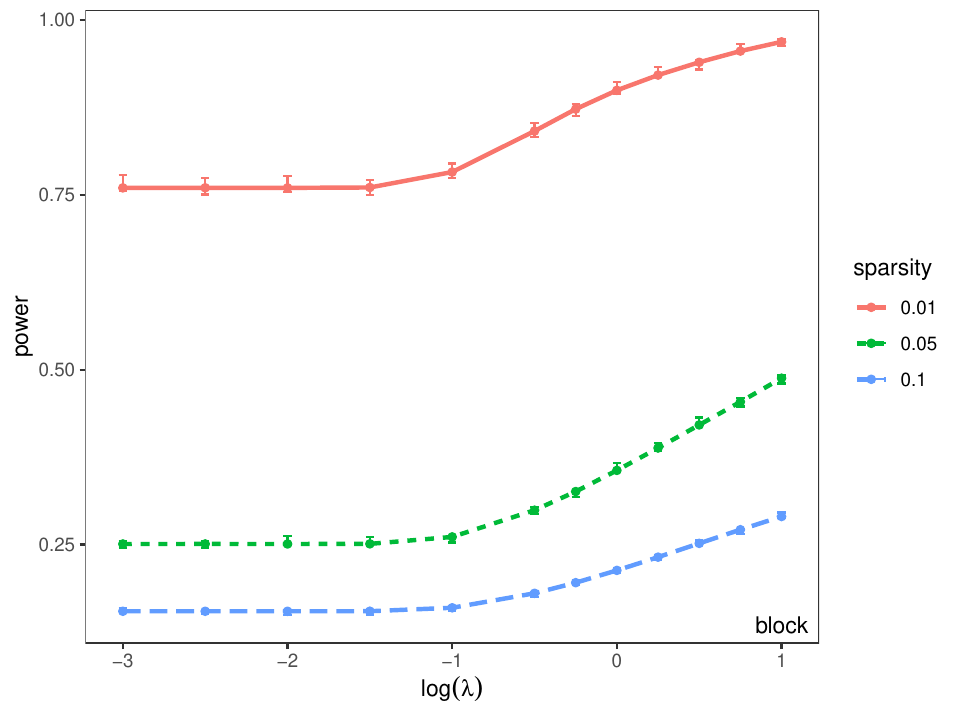}
\includegraphics[width = 0.45\textwidth]{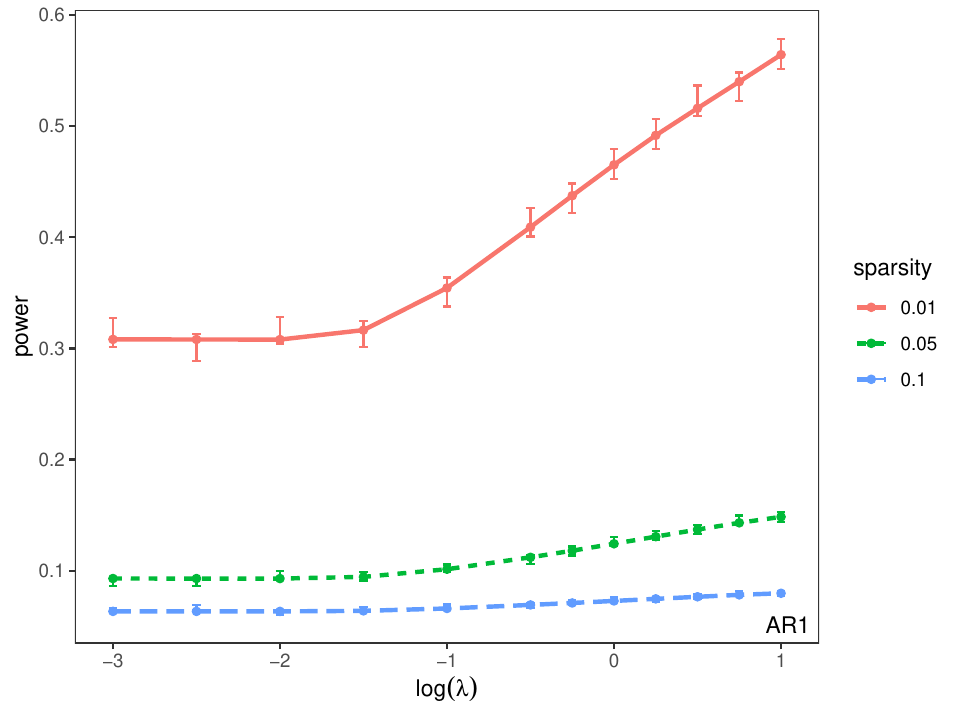}
\includegraphics[width = 0.45\textwidth]{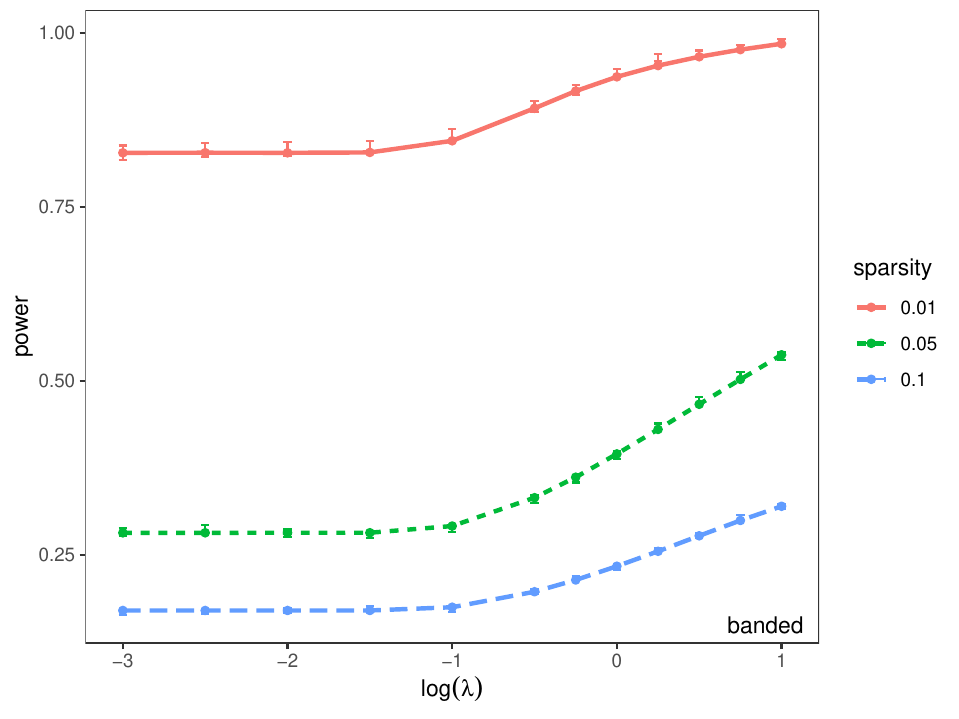}
\end{center}
\caption{Empirical and asymptotic powers as a function of $\log\lambda$ for four different correlation structures: IID (top-left), block (top-right), AR(1) (bottom-left), and banded (bottom-right). The lines are the asymptotic powers at different sparsity levels $s = 0.01, 0.05, 0.1$. The error bars are the 95\% confidence intervals for the mean powers based on 500 replicates.}
\label{fig:5-power}
\end{figure}

Next, we assess variable selection performance using power, defined as the probability of correctly identifying a significant variable when its true coefficient is nonzero. The asymptotic power is derived from the results in Section~\ref{section:main-results}, while the empirical power is estimated as the proportion of nonzero components whose 95\% confidence intervals do not include zero.
Figure~\ref{fig:5-power} shows the asymptotic power as lines and the 95\% confidence intervals of the mean empirical powers based on 500 replicates as error bars. The close agreement between the asymptotic and empirical powers validates the theoretical results. The power increases as the true underlying model becomes sparser, because the effective dimension of the model is reduced. The power also increases with $\lambda$, because a larger $\lambda$ shrinks more insignificant variables to zero, making the significant variables easier to detect.

\begin{figure}[htb]
\begin{center}
\includegraphics[width = 0.45\textwidth]{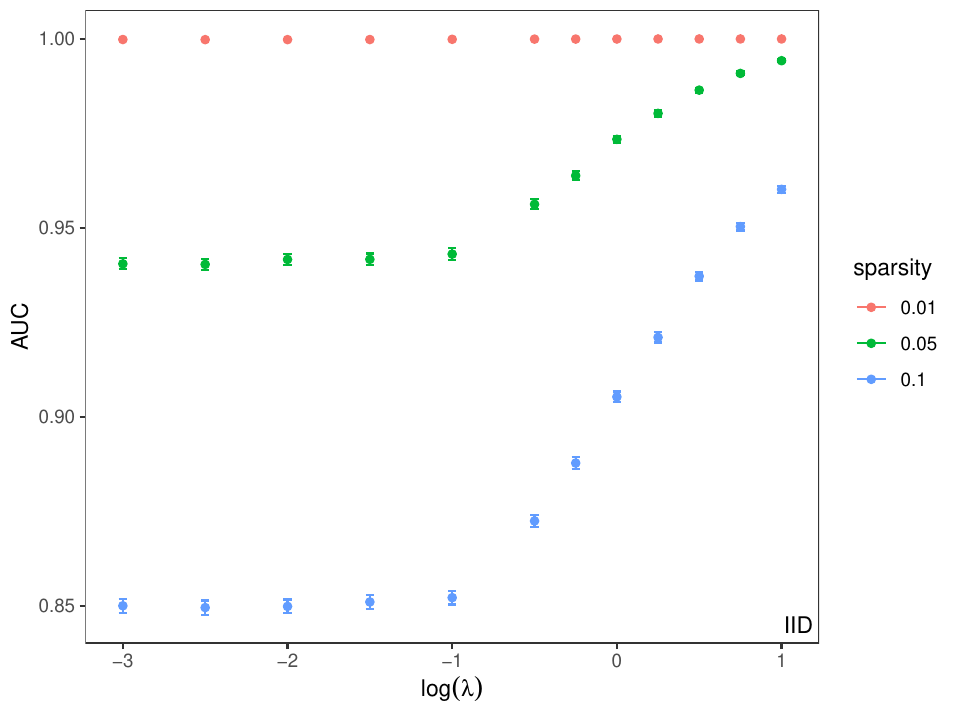}
\includegraphics[width = 0.45\textwidth]{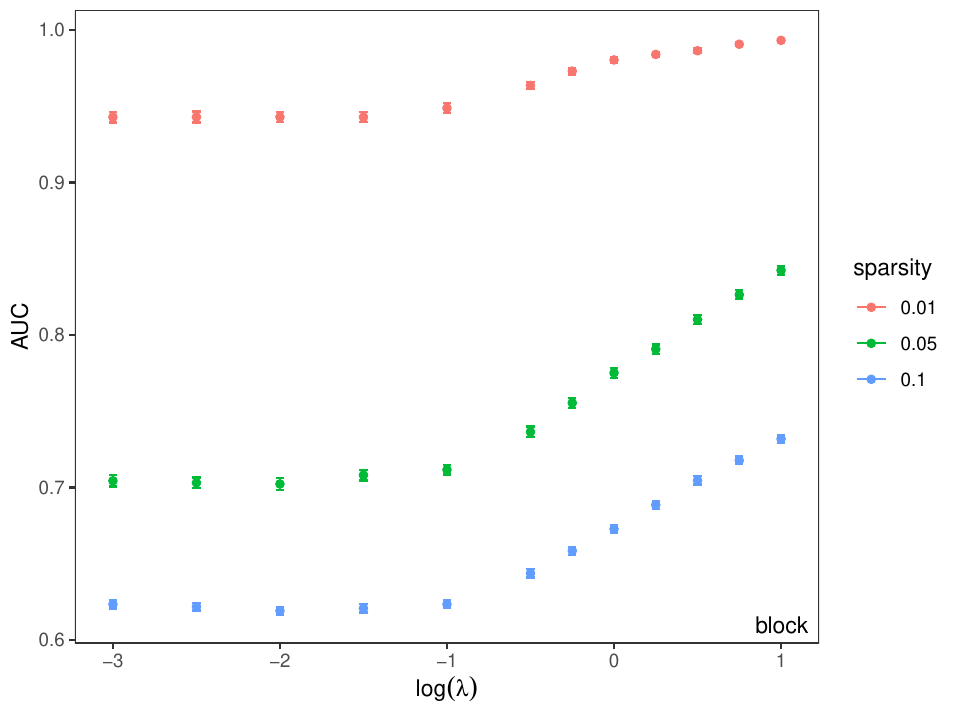}
\includegraphics[width = 0.45\textwidth]{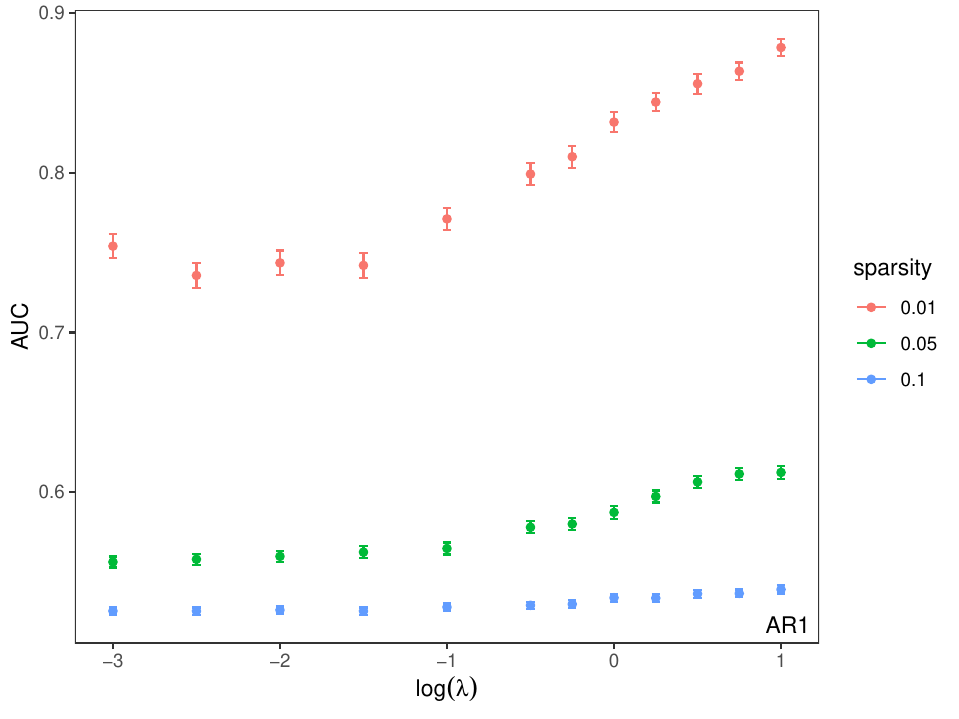}
\includegraphics[width = 0.45\textwidth]{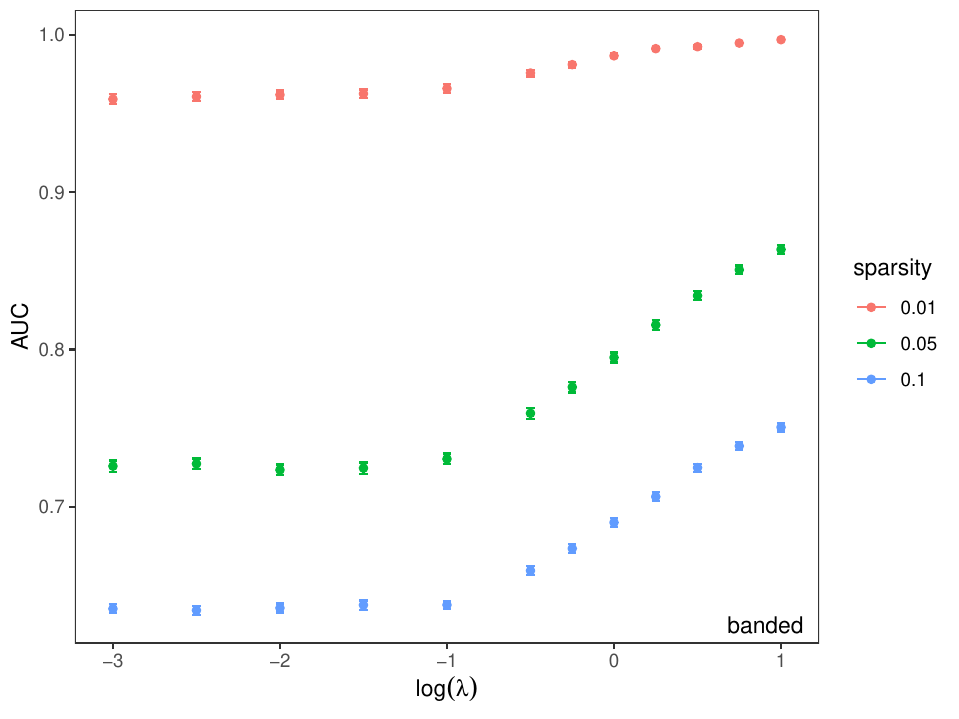}
\end{center}
\caption{AUC as a function of $\log\lambda$ for four different correlation structures: IID (top-left), block (top-right), AR(1) (bottom-left), and banded (bottom-right). The error bars are the 95\% confidence intervals for the mean AUC based on 500 replicates at different sparsity levels $s = 0.01, 0.05, 0.1$.}
\label{fig:7-auc}
\end{figure}

The area under the curve (AUC) is a common measure of variable selection performance. For each component, the $p$-value corresponding to the null hypothesis that the component is insignificant can be computed using the results in Section~\ref{section:main-results}. These $p$-values allow us to construct the ROC curve and compute the AUC. Figure~\ref{fig:7-auc} presents the 95\% confidence intervals for the AUC under different correlation structures and sparsity levels. The performance of variable selection improves as the true underlying model becomes sparser. The relationship between AUC and $\lambda$ is similar to that between power and $\lambda$ in Figure~\ref{fig:5-power}.

\subsection{Real data analysis}


The human breast cancer dataset used in this section is from the MicroArray Quality Control (MAQC) II project \citep{popovici2010effect, maqc2010microarray}. The data are publicly available through the Gene Expression Omnibus (GEO) database (\verb"https://www.ncbi.nlm.nih.gov/geo/") under accession number GSE20194.
The study includes stage I--III breast cancer patients, of whom 164 have positive clinical estrogen receptor status and 114 have negative status. Thus the observed classes are moderately imbalanced. Statistical-physics analyses of high-dimensional linear classification under class imbalance, including \citet{Loffredo2024RestoringBalance}, show that class proportions can affect classification performance. Because the theoretical analysis in this paper assumes a balanced Gaussian mixture, we regard the real-data analysis as an illustrative application of the fitted framework rather than as a validation of a theory that explicitly models class imbalance.
For each subject, gene expression measurements are available for 22,283 genes.

We first perform two-sample $t$-tests for each gene separately and select the top $p = 1000$ genes with the smallest $p$-values for further analysis. This preliminary screening is used to reduce the computational dimension and should be viewed as a data-reduction step rather than part of the formal inferential guarantee developed in Sections~\ref{section:main-results}--\ref{section:implementation}. In particular, the reported post-screening $p$-values are interpreted as illustrative rather than as unconditional end-to-end inference over all 22,283 genes. A sample-splitting analysis would provide a more stringent inferential interpretation and is an important direction for future work.

The gene expression data are then recentered so that the mean expression levels of the two groups are $\pm\bmu$, respectively.
The covariance matrix is estimated using the principal orthogonal complement thresholding (POET) method  \citep{fan2013large}. We estimate the covariance matrices for the two groups separately, and then compute their average weighted by the respective sample sizes. Set $\alpha = n/p$, where $n = 164 + 114$ is the total sample size.
For each given $\lambda$, we compute the saddlepoint parameters and the debiased estimates. The resulting coordinatewise $p$-values for testing $H_0:w_j=0$ are adjusted for multiple comparisons using the false discovery rate procedure \citep{benjamini1995controlling}. Because the preliminary screening is data-dependent, these results are intended to illustrate the distinction between variables selected by the sparse SVM and variables supported by the proposed inferential procedure rather than to claim unconditional genome-wide error-rate control.

\begin{figure}[hbt]
\centering
\includegraphics[width = 0.5\textwidth]{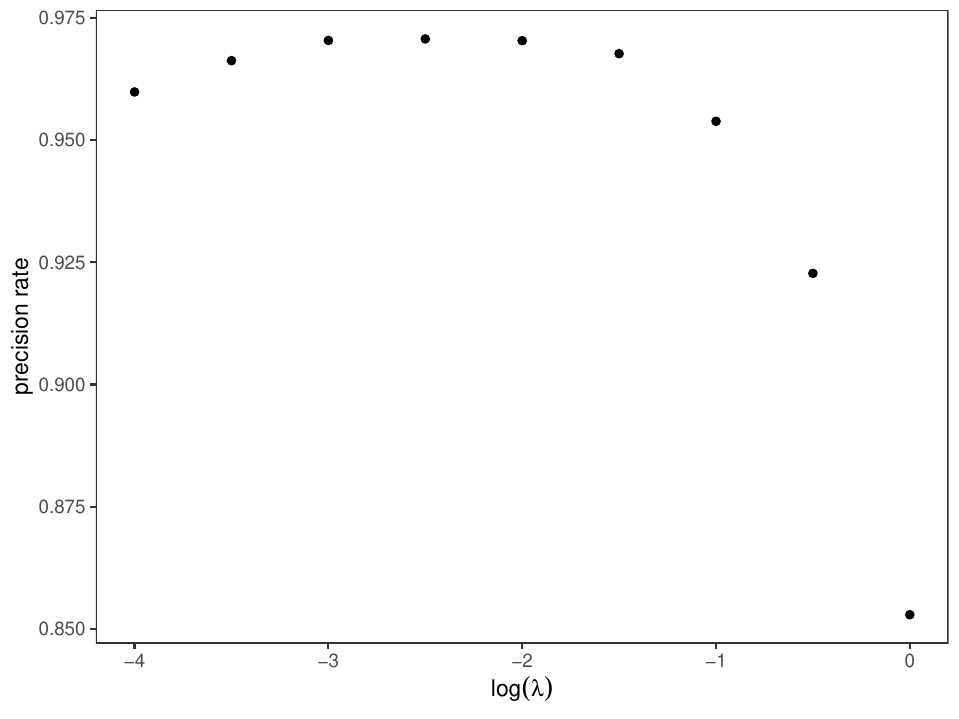}
\caption{Asymptotic classification accuracies as a function of $\log(\lambda)$ for the breast cancer dataset.}
\label{fig:cancer-precision}
\end{figure}

Figure~\ref{fig:cancer-precision} displays the asymptotic classification accuracy as a function of $\log(\lambda)$. The classification accuracy first increases and then decreases as $\log(\lambda)$ increases. The optimal classification accuracy is achieved at $\log(\lambda) = -2.5$, at which point the procedure identifies 60 significant genes.
For comparison, the lasso estimate $\hat\bw$ has 99 nonzero components.

\section{Conclusion}
\label{section:conclusion}

We have developed a high-dimensional inferential framework for the $L_1$-penalized SVM in the proportional asymptotic regime where $n,p\to\infty$ with $n/p\to\alpha$. The central mathematical difficulty is the nonsmooth hinge loss. Rather than treating this difficulty as a minor technical modification of a smooth-loss theory, we exploit the exact linear-program structure of the sparse SVM: its dual variables identify a valid hinge-loss subgradient and therefore provide the quantities required by the debiasing correction. This yields the directly computable representation
\[
\bar \bw
=
\hat \bw
+
\frac{1}{\zeta}\bSigma^{-1}
\sum_{i=1}^{n}\nu_i y_i\bx_i.
\]
The resulting Gaussian approximation provides coordinatewise confidence intervals and hypothesis tests and offers a principled basis for multiple-testing and variable-selection procedures.

The simulation studies show that the asymptotic characterization accurately describes classification performance and inferential calibration across several covariance structures, including strongly correlated designs. The breast cancer application illustrates the practical distinction between the variables selected by the raw sparse SVM and those supported by the proposed inferential procedure. The current real-data analysis uses preliminary feature screening for computational reduction; consequently, its post-screening inferential results should be interpreted as illustrative rather than as unconditional genome-wide inference.

Several directions remain open. First, the present theory is restricted to balanced, linearly separable data; extending it to class-imbalanced designs \citep{Loffredo2024RestoringBalance} or to the nonseparable perceptron regime \citep{Franz2019CriticalJammed} is a natural next step. Second, the theoretical framework is based on a replica-symmetric characterization and does not provide a rigorous proof of the RS ansatz or its stability; establishing such results, or identifying conditions under which the RS solution can be justified rigorously, is an important theoretical problem. Third, the framework assumes a Gaussian-mixture model for the class-conditional distributions, and extending the theory to broader distributional classes would increase its applicability. Fourth, the current analysis is restricted to the $L_1$ penalty; extending the proportional-asymptotic theory to nonconvex penalties such as SCAD \citep{fan2001variable} or MCP could combine reduced regularization bias with formal inferential guarantees. Fifth, a theoretically justified, data-driven procedure for selecting the tuning parameter $\lambda$ remains an important problem. Finally, developing sample-splitting or other selection-aware procedures for the preliminary screening step in high-dimensional applications would provide a cleaner end-to-end inferential framework.

\bibsep 5pt
\bibliographystyle{chicago}
\bibliography{references}

\end{document}